\documentclass[11pt,a4paper]{article}
\usepackage[nohyperref]{emnlp-ijcnlp-2019}
\usepackage{times}
\usepackage{latexsym}
 \usepackage{graphicx}

\usepackage{url}
\aclfinalcopy 

\title{On the use of BERT for Neural Machine Translation}

\author{St\'ephane Clinchant \\
  NAVER LABS Europe, France \\
  {\tt \small stephane.clinchant@naverlabs.com}  \\\And
  Kweon Woo Jung \\
  NAVER Corp., \\
  South Korea\\
  {\tt \small kweonwoo.jung@navercorp.com}\\\And
  Vassilina Nikoulina \\
  NAVER LABS Europe, France \\
  {\tt \small vassilina.nikoulina@naverlabs.com} \\}

\date{}

\begin{document}
\maketitle
\begin{abstract}
Exploiting large pretrained models for various NMT tasks have gained a lot of visibility recently. In this work we study how BERT pretrained models could be exploited for supervised Neural Machine Translation.  We compare various ways to integrate pretrained BERT model with NMT model and study the impact of the monolingual data used for BERT training on the final translation quality. We use WMT-14 English-German, IWSLT15 English-German and IWSLT14 English-Russian datasets for these experiments. In addition to standard task test set evaluation, we perform evaluation on out-of-domain test sets and noise injected test sets, in order to assess how BERT pretrained representations affect model robustness. 
\end{abstract}

\section{Introduction}

Pretrained Language Models (LM) such as ELMO and BERT \cite{ELMO,Bert2018} have turned out to significantly improve the quality of several Natural Language Processing (NLP) tasks by transferring the prior knowledge learned from data-rich monolingual corpora to data-poor NLP tasks such as question answering, bio-medical information extraction and standard benchmarks \cite{glue, biobert}. In addition, it was shown that these representations contain syntactic and semantic information in different layers of the network \cite{tenney2019bert}. Therefore, using such pretrained LMs for Neural Machine Translation (NMT) is appealing, and has been recently tried by several people \cite{XLM2019, EdunovEtAl2019, song2019mass}. 

Unfortunately, the results of the above-mentioned works are not directly comparable to each other as they used different methods, datasets and tasks. Furthermore,  pretrained LMs have mostly shown improvements  in low-resource or unsupervised NMT settings, and has been little studied in standard supervised scenario with reasonable amount of data available. 

Current state of the art NMT models rely on the Transformer model \cite{transformer}, a feed-forward network relying on attention mechanism, which has surpassed prior state of the art architecture based on recurrent neural nets \cite{bahdanau_rnn, sutskever_s2s}. 
Beyond machine translation, the transformer models have been reused to learn bi-directional language models on large text corpora. The BERT model \cite{Bert2018} consists in a transformer model aiming at solving a masked language modelling task, namely correctly predicting a masked word from its context, and a next sentence prediction task  to decide whether two sentences are consecutive or not. 
In this work, we  study how pretrained BERT models can be exploited for \textit{transformer-based} NMT, thus exploiting the fact that they rely on the same architecture. 

The objective of this work is twofold. On one hand, we wish to perform systematic comparisons of different BERT+NMT architectures for standard supervised NMT. In addition, we argue that the benefits of using pretrained representations has been overlooked in previous studies and should be assessed beyond BLEU scores on in-domain datasets. In fact, LMs trained on huge datasets have the potentials of being more robust in general and improve the performance for domain adaptation in MT.

In this study, we compare different ways to train and reuse BERT for NMT. For instance, we show that BERT can be trained only with a masked LM task on the NMT source corpora and yield significant improvement over the baseline. In addition, the models robustness is analyzed thanks to synthetic noise.

The paper is organized as follows. In section \ref{sec:soa}, we review relevant state of the art. Section \ref{sec:exp_study} enumerates different models we experiment with. Finally section \ref{sec:exp_setup} and \ref{sec:iwslt} present our results before discussing the main contributions of this work in section \ref{sec:discussion}.

\section{Related Works}\label{sec:soa}


The seminal work of \cite{Bengio:2003_LM,CollobertWeston_2005} were one of the first to show that neural nets could learn word representations useful in a variety of NLP tasks, paving the way for the word embedding era thanks to {\it{word2vec}} \cite{word2vec} and its variants \cite{glove,Levy:2014}.

With the recent advances and boost in performance of neural nets, ELMO \cite{ELMO} employed a Bi-LSTM network for language modelling and proposed to combine the different network layers to obtain effective word representations. Shortly after the publication of ELMO, the BERT model \cite{Bert2018} was shown to have outstanding performance in various NLP tasks. Furthermore, the BERT model was refined in 
\cite{cloze2019fair} where the transformer self-attention mechanism is replaced by two directional self-attention blocks:
a left-to-right and right-to-left blocks are combined to predict the masked tokens.

With respect to NMT, backtranslation \cite{sennrich-bt} is up to now one of the most effective ways to exploit large monolingual data. However, backtranslation has the drawback of being only applicable for target language data augmentation, while pretrained LMs can be used both for source and target language (independently \cite{EdunovEtAl2019} or jointly \cite{XLM2019, song2019mass}). 

\newcite{XLM2019} initializes the entire encoder and decoder with a pretrained MaskLM  or Crosslingual MaskLM language models trained on multilingual corpora. Such initialization proved to be beneficial for unsupervised machine translation, but also for English-Romanian supervised MT, bringing additional improvements over standard backtranslation  with MLM initialization.   

\newcite{EdunovEtAl2019} uses ELMO \cite{ELMO} language model to set the word embeddings layer in NMT model. In addition, the ELMO embedding are compared with the cloze-style BERT \cite{cloze2019fair} ones. The embedding network parameters are then either fixed, or fine-tuned. This work shows improvements on English-German and English-Turkish translation tasks when using pretrained language model for source word embedding initialization. However, the results are less clear when reusing embedding on the target language side. 

Futhermore, \newcite{song2019mass} goes one step further and proposes Masked Sequence-to-Sequence pretraining method. Rather than masking a single token, it masks a sequence of token in the encoder and recovers them in the decoder. This model has shown new state of the art for unsupervised machine translation. 

Our work is an attempt to perform systematic comparison on some of the aforementioned architectures that incorporate pretrained LM in NMT model, concentrating on BERT pretrained LM representations applied on supervised machine translation. However, we restrict ourselves to encoder part only, and leave the decoder initialization for future work.

Regarding robustness, several recent studies \cite{karpukhin_robustness,singh_robustness} have tackled robustness issues with data augmentation. In this work,  we study whether the robustness problem can be addressed at the model level rather than at data level.  \newcite{adversarial_perturbations} address robustness problem with generative adversarial networks. This method, as well as data augmentation methods are complementary to our work and we believe that they address different issues of robustness. 

%
\section{Methods}
\label{sec:exp_study}

Typical NMT model adopts the encoder-decoder architecture where the encoder forms contextualized word embedding from a source sentence and the decoder generates a target translation from left to right. 

Pretrained LM, namely BERT, can inject prior knowledge on the encoder part of NMT, providing rich contextualized word embedding learned from large monolingual corpus. Moreover, pretrained LMs can be trained once, and reused for different language pairs\footnote{As opposed to backtranslation techniques which requires full NMT model retraining}. 

In this study, we focus on reusing BERT models for the NMT encoder\footnote{Similar  approach can be applied on the target language but we leave it for future work.}. We will compare the following models:
\begin{itemize}
    \item \textbf{Baseline:}. A \textit{transformer-big} model with shared decoder input-output embedding parameters. 
    \item {\bf Embedding (Emb): } The baseline model
where the embedding layer is replaced by the BERT parameters (thus having $6+6$ encoder layers). The model is then fine tuned similar to the ELMO setting from \cite{EdunovEtAl2019}
    \item {\bf Fine-Tuning (FT):}  The baseline model with the encoder initialized by the BERT parameters as in \newcite{XLM2019}
    \item {\bf Freeze:}  The baseline model with the encoder initialized by the BERT parameters and frozen. This means that the whole encoder has been trained in purely monolingual settings, and only parameters responsible for the translation belong to the attention and decoder models.  
\end{itemize}

We exploit the fact that BERT uses the same architecture as NMT encoder which allows us to initialize NMT encoder with BERT pretrained parameters. BERT pretraining has two advantages over NMT training:
\begin{itemize}
    \item it solves a simpler (monolingual) task of `source sentence encoding', compared to NMT (bilingual task) which has to `encode source sentence information', and `translate into a target language'.
    \item it has a possibility to exploit much larger data, while NMT encoder is limited to source side of parallel corpus only.
\end{itemize}

Even though the role of NMT encoder may go beyond source sentence encoding (nothing prevents the model from encoding `translation related' information at the encoder level), better initialization of encoder with BERT pretrained LM allows for faster NMT learning. 
Comparing settings where we freeze BERT parameters against fine-tuning BERT allows to shed some light on the capacity of the encoder/decoder model to learn `translation-related' information. 

Moreover, since the BERT models are trained to predict missing tokens from their context, their representations may also be more robust to missing tokens or noisy inputs. We perform extensive robustness study at section \ref{sec:exp_setup} verifying this hypothesis.

Finally, language models trained on huge datasets have the potentials of being more robust in general and improve the performance for domain adaptation in MT. We therefore compare BERT models trained on different datasets, and perform evaluation on related test sets in order to assess the capacity of pretrained LMs on domain adaptation.  



\section{WMT experiments}
\label{sec:exp_setup}

\subsection{Preprocessing} We learn BPE \cite{sennrich-bpe} model with 32K split operations on the concatenation of Wiki and News corpus. This model is used both for Pre-trained LM subwords splitting and NMT source (English) side subwords splitting. German side of NMT has been processed with 32K BPE model learnt on target part of parallel corpus only. Please note, that this is different from standard settings for WMT En-De experiments, which usually uses joint BPE learning and shared source-target embeddings.
We do not adopt standard settings since it contradicts our original motivation for using  pre-trained LM: English LM is learnt once and reused for different language pairs.     

\subsection{Training}

\paragraph{BERT}
For pretraining BERT models, we use three different monolingual corpora of different sizes and different domains. 
Table \ref{tab:mono_data} summarizes the statistics of these three monolingual corpora. 
\begin{itemize}
    \item {\it{NMT-src}}: source part of our parallel corpus that is used for NMT model training.
    \item {\it{Wiki}}: English wikipedia dump
    \item {\it{News}}: concatenation of 70M samples from "News Discussion", "News Crawl" and "Common Crawl" English monolingual datasets distributed by WMT-2019 shared task\footnote{http://www.statmt.org/wmt19/translation-task.html}. This resulted in total 210M samples.
    
\end{itemize}

The motivation of using NMT-src is to test whether the resulting NMT model is more robust after having being trained on the source corpora. The Wiki corpora is bigger than the NMT-src but could be classified as out-of-domain compared to news dataset. Finally, the news dataset is the biggest one and consists mostly of in-domain data.
 
In all of our experiments, we only consider using the masked LM task for BERT as the next sentence prediction tasks put restrictions on possible data to use. We closely follow the masked LM task described in \cite{Bert2018} with few adjustments optimized for downstream NMT training. We use frequency based sampling \cite{XLM2019} in choosing 15\% of tokens to mask, instead of uniformly sampling. Instead of MASK token we used UNK token hoping that thus trained model will learn certain representation for unknowns that could be exploited by NMT model.  Warm-up learning scheme described in \cite{transformer} results in faster convergence than linear decaying learning rate. The batch size of 64000 tokens per batch is used, with maximum token length of 250, half the original value, as we input single sentence only. We do not use [CLS] token in the encoder side, as attention mechanism in NMT task can extract necessary information from token-level representations. The BERT model is equivalent to the encoder side of Transformer Big model. We train BERT model up to 200k iterations until the accuracy for masked LM on development saturates.

\begin{table}[]
    \centering
    \begin{tabular}{r|r|r}
         & Lines & Tokens \\\hline
         NMT-src & 4.5M & 104M \\
         Wiki &  72M & 2086M \\
         News & 210M & 3657M\\\hline
    \end{tabular}
    \caption{Monolingual (English) training data}
    \label{tab:mono_data}
\end{table}

\paragraph{NMT}
For NMT system training, we use WMT-14 English-German dataset.

We use \textit{Transformer-Big} as our baseline model. We share input embedding and output embedding parameters just before softmax on the decoder side. Warm up learning scheme is used with warm-up steps of 4000. We use batch size of 32000 tokens per batch. Dropout of 0.3 is applied to residual connections, and no dropout is applied in attention layers.  We decode with beam size 4 with length penalty described in \newcite{gnmt16}. We conduct model selection with perplexity on development set. We average 5 checkpoints around lowest perplexity.

\subsection{Evaluation}
\label{sec:eval}
We believe that the impact of pretrained LM in NMT model can not be measured by BLEU performance on in-domain test set only. Therefore we introduce additional evaluation that allows to measure the impact of LM pretraining on different out-of-domain tests. We also propose an evaluation procedure to evaluate the robustness to various types of noise for our models. 

\paragraph{Domain} Besides standard WMT-14 news test set, models are evaluated on additional test sets given by Table \ref{tab:test_sets_domain}. We include two in-domain (news) test sets, as well as additional out-of-domain test sets described in Table \ref{tab:test_sets_domain}. 
\begin{table}[t]
    \centering
    \begin{tabular}{l|c|c}
         & Lines & Tok/line (en/de)\\\hline
         news14 & 3003 & 19.7/18.3\\
         news18 & 2997 & 19.5/18.3\\
         iwslt15 & 2385 & 16.4/15.4\\
         OpenSub & 5000 & 6.3/5.5\\
         KDE & 5000 & 8/7.7\\
         wiki & 5000 & 17.7/15.5\\
    \end{tabular}
    \caption{ In/Out of Domain test sets. news14 and news18 are test sets from WMT-14 and WMT-18 news translation shared task. iwslt: test set from IWSLT-15 MT Track\footnote{https://sites.google.com/site/iwsltevaluation2015/mt-track}. Wiki is randomly 5K sampled from parallel Wikipedia distributed by OPUS\footnote{http://opus.nlpl.eu/}, OpenSub, KDE and Wiki are randomly 5K sampled from parallel Wikipedia, Open Subtitles and KDE corpora distributed by OPUS\footnote{http://opus.nlpl.eu/}
    }
    \label{tab:test_sets_domain}
\end{table}

\paragraph{Noise robustness.} 
For robustness evaluation,  we introduce different type of noise to the standard \textit{news14} test set:

{\bf Typos:} Similar to \newcite{karpukhin_robustness}, we add synthetic noise to the test set 
 by randomly (1) swapping characters (chswap), (2) randomly inserting or deleting characters (chrand), (3) uppercasing words (\textit{up}). These test sets translations are evaluated against the golden \textit{news14} reference. 
 
{\bf Unk}: An unknown character is introduced at the beginning (noted \textit{UNK.S}) or at the end of the sentence (noted \textit{UNK.E}) before a punctuation symbol if any (this unknown character could be thought as as an unknown emoji, a character in different script, a rare unicode character). This token is introduced both for source and target sentence, and the evaluation is performed with the augmented-reference. 

Intuitively, we expect the model to simply copy UNK token and proceed to the remaining tokens. Interestingly, this simple test seems to produce poor translations, therefore puzzling the  attention and decoding process a lot. Table \ref{table:unk_example} gives an example of such translations for baseline model\footnote{Output for (UNK.S+src) input is not an error, the model does produces an English sentence!}.

Since the tasks are correlated, a better model might be better on noisy test sets as it behaves better in general. If we want to test that some models are indeed better, we need to disentangle this effect and show that the gain in performance is not just a random effect. A proper way would be to compute the BLEU correlation between the original test set and the noisy versions but it would require a larger set of models for an accurate correlation estimation.  

 {\bf $\Delta$(chrF) :} We propose to look at the \textit{distribution of the difference of sentence charf between the noisy test set and the original test set}. Indeed, looking at BLEU delta may not provide enough information since it is corpus-level metric. Ideally, we would like to measure a number of sentences or a margin for which we observe an `important decrease' in translation quality. According to \newcite{ma-bojar-graham:2018:WMT, bojar-graham-kamran:2017:WMT}, sentence level chrF  achieves good correlation with human judgements for En-De news translations. 
 
 More formally, let $s$ be a sentence from the standard news14 test set, $n$ a  noise operation, $m$ a translation model and $r$ the reference sentence\footnote{In the case of UNK transformation, the reference is changed but we omit that to simplify the notation.}:
 \begin{eqnarray}
     \Delta(\textrm{chrF)}(m,n,s)&=& \textrm{chrF}(m(n(s)),r) - \nonumber \\ 
                                  & & \textrm{chrF}(m(s),r)
 \end{eqnarray}
 In the analysis, we will report the distribution of {\bf $\Delta$(chrF)} and    its mean value as a summary. If a model is good at dealing with noise, then the produced sentence will be similar to the one produced by the noise-free input sentence. Therefore, the $\Delta(\textrm{chrF)}$ will be closer to zero.

\begin{table*}[]
    \centering
    \begin{tabular}{l|l}
        source sentence & "In home cooking, there is much to be discovered - with a few minor \\ 
         & tweaks you can achieve good, if not sometimes better results," said Proctor.\\ 
         \hline
        translation(src) & "Beim Kochen zu Hause gibt es viel zu entdecken - mit ein paar kleinen \\
        & Änderungen kann man gute, wenn nicht sogar manchmal bessere \\
        & Ergebnisse erzielen", sagte Proktor.\\
        \hline
         translation(UNK.S + src) & $\bullet $ "In home cooking, there is much to be discovered - with a few minor \\
         & tweaks you can achieve good, if not sometimes better results", sagte Proktor. \\
    \end{tabular}
    \caption{Example of a poor translation when adding unknown token to source sentences (translation done with a baseline transformer model)}
    \label{table:unk_example}
    \label{tab:my_label}
\end{table*}

\subsection{Results}
Table \ref{tab:diff_nmt_bert_arch} presents the results of our experiments. As expected, freezing the encoder with BERT parameters lead to a significant decrease in translation quality. However, other BERT+NMT architectures mostly improve over the baseline both on in-domain and out-of-domain test sets. We conclude, that the information encoded by BERT is useful but not sufficient to perform the translation task. We believe, that the role of the NMT encoder is to encode both information specific to source sentence, but also information specific to the target sentence (which is missing in BERT training). 

Next, we observe that even NMTSrc.FT (NMT encoder is initialized with BERT trained on source part of parallel corpus) improves over the baseline. Note that this model uses the same amount of data as the baseline. BERT task is simpler compared to the task of the NMT encoder, but it is still related, thus BERT pretraining allows for a better initialization point for NMT model. 

When using more data for BERT training (Wiki.FT and News.FT), we gain even more improvements over the baseline. 

Finally, we observe comparable results for News.Emb and News.FT (the difference in BLEU doesn't exceed 0.3 points, being higher for News.FT on in-domain tests, and News.Emb for out-of-domain tests). Although News.FT configuration keeps the size of the model same as standard NMT system, News.Emb adds BERT parameters to NMT parameters which doubles the size of NMT encoder. Additional encoder layers introduced in News.Emb does not add significant value.
\begin{table*}[t!]
    \centering
    \begin{tabular}{l| cccccc }
          & news14 & news18 & iwslt15 & wiki & kde & OpenSub  \\\hline 
         Baseline & 27.3 & 39.5 & 28.9 & 17.6 & 18.1 &15.3 \\\hline
         NMTsrc.FT & 27.7 & 40.1 & 28.7 & 18.3 & 18.4 & 15.3 \\
         Wiki.FT & 27.7 & \textbf{40.6} & 28.7 & 18.4&  \textbf{19.0} & 15.4 \\
         News.FT & \textbf{27.9} & 40.2 & 29.1 & 18.8 & 17.9 & 15.7 \\\hline
         News.Emb & 27.7 & 39.9 & \textbf{29.3} & \textbf{18.9} & 18.2 & \textbf{16.0} \\\hline
         News.Freeze & 23.6 & 35.5 & 26.5 & 15.0 & 15.1 & 13.8 \\\hline

    \end{tabular}
    \caption{FT: initialize NMT encoder with BERT and finetune; Freeze: fix NMT encoder parameters to BERT parameters; Emb: fix encoder embeddding layer with BERT contextual word embeddings. }
    \label{tab:diff_nmt_bert_arch}
\end{table*}

\subsection{Robustness analysis}\label{sec:robust_res}

\begin{table*}[]
    \centering
    \begin{tabular}{l|c|cc|ccc}
         Models  & news14 & +UNK.S & +UNK.E &+chswap &  +chrand & +up  \\\hline
         Baseline & 27.3 & 24.8 & 24.4 & 24.2 & 24.7 & 23.5 \\
         {NMTsrc.FT} & 27.7 & 24.9 & 22.9 &  24.4 &25.2 & \textbf{24.5} \\ 
         {Wiki.FT}& 27.7 & \textbf{25.8} & \textbf{24.9} &  24.4 & 24.9 &  24.4 \\
    
         {News.FT} & \textbf{27.9} & 24.9 & \textbf{24.9} & 24.5 & \textbf{25.3} & \textbf{24.5} \\
         {News.Emb} & 27.7 & 24.7 & 24.8  & \textbf{ 24.6} & \textbf{25.3} & 24.2 \\\hline
                \end{tabular}
    \caption{Robustness tests. BLEU scores for clean and 'noisified' (with different noise type) \textit{news14} testset.}
    \label{tab:eval_robust1}
\end{table*}

Table \ref{tab:eval_robust1} reports BLEU scores for the noisy test sets (described in section \ref{sec:eval}). As expected, we observe an important drop in BLEU scores due to the introduced noise.
We observe that most pretrained BERT models have better BLEU scores compared to baseline for all type of noise (except NMTSrc.FT which suffers more from unknown token introduction in the end of the sentence compared to the Baseline). 
 However, these results are not enough to conclude, whether higher BLEU scores of BERT-augmented models are due to better robustness, or simply because these models are slightly better than the baseline in general.

\begin{figure}
    \centering
    \includegraphics[width=0.5\textwidth]{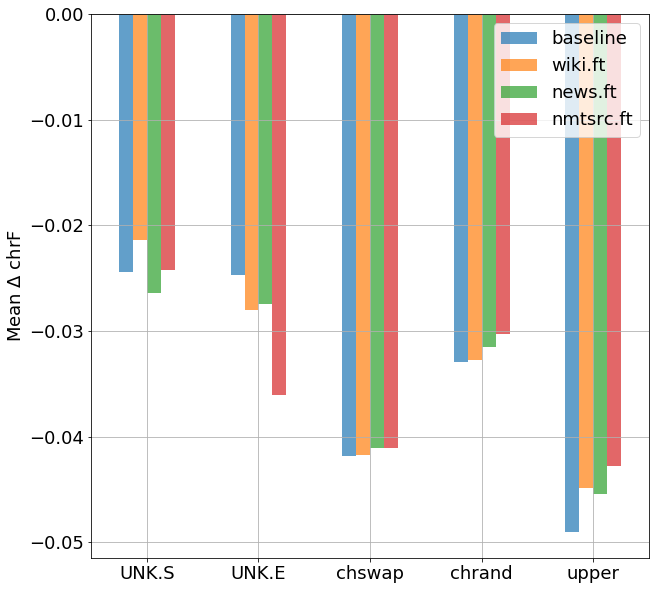}
    \caption{ Mean {\bf $\Delta$(chrF) } for several noisy test set and models. For the UNK test, the BERT models are similar or worst than the basline. For the chrand, chswap, upper, the BERT models are slightly better.}
    \label{fig:charf_mean}
\end{figure}

This is why figure \ref{fig:charf_mean} reports the mean $\Delta$(chrF) for several models. $\Delta$(chrF) scores for UNK tests show that BERT models are not better than expected. However, for chswap, chrand, upper, the BERT models have a slightly lower $\Delta$(chrF). Based on these results, we conclude that pretraining the encoder with a masked LM task does not really bring improvement in terms of robustness to unknowns. It seems that BERT does yield improvement for NMT as a better initialization for NMT encoders but the full potential of masked LM task is not fully exploited for NMT. 

\section{IWSLT experiments}\label{sec:iwslt}
In order to explore the potential of masked LM encoder pretraining for NMT in lower resource settings, we train NMT models on English-German IWSLT 2015\footnote{https://sites.google.com/site/iwsltevaluation2015/mt-track} and English-Russian IWSLT 2014\footnote{https://sites.google.com/site/iwsltevaluation2014/mt-track} MT track datasets. These are pretty small datasets (compared to previous experiments) which contain around 200K parallel sentences each.

\subsection{Experimental settings}
In these experiments we (1) reuse pretrained BERT models from previous experiments or (2) train IWSLT BERT model. IWSLT BERT model is trained on the concatenation of all the data available at IWSLT 2014-2018 campaigns. After filtering out all the duplicates it contains around 780K sentences and 13.8M tokens. 

We considered various settings for IWSLT baseline. First, for source side of the dataset, we took 10K BPE merge operations, where BPE model was trained  (1) either on the source side of NMT data only, or (2) on all monolingual English IWSLT data. Target side BPE uses 10K merge operations trained on the target side of the NMT dataset in all the IWSLST experiments. In our first set of experiments, BPE model learnt on source data only lead to similar translation performance as BPE model learnt on all IWSLT English data. Therefore, in what follows we report results only for the latter (referred as $bpe10k$).

NMT model training  on IWSLT datasets with Transformer Big architecture on IWSLT data has diverged both for en-de and en-ru dataset. Therefore we use Transformer Base ($tbase$) architecture as a baseline model for these experiments. IWSLT BERT model is also based on $tbase$ architecture described in \newcite{transformer} and for the rest follows same training procedure as 
described in the section \ref{sec:exp_setup}. 

In order to explore the potential of single pretrained model for all language pairs/datasets we try to reuse Wiki and News pretrained BERT models from previous experiments for encoder initialization of NMT model.
However, in the previous experiments, our pretrained BERT models used 32K BPE vocabulary and Transformer Big ($tbig$) architecture which means that we have to reuse the same settings for the encoder trained on IWSLT dataset. It has been shown by \newcite{prudentBPE}, these are not optimal settings for IWSLT training because it leads to too many parameters for the amount of data available. Therefore, in order to reduce the amount of the parameters of the model, we  also consider the case where we reduce the amount of the decoder layers from 6 to 3 ($tbig.dec3$). 

\subsection{Results}

\begin{table}[t!]
    \centering
    \begin{tabular}{l| cc }

          & en-de & en-ru   \\\hline 
          & \multicolumn{2}{c}{Baseline}\\\hline
         $tbase.bpe10k$ & 25.9 &  9.6 \\
         $tbase.dec3.bpe10k$ & 26.4 & 16.3  \\\hline
         & \multicolumn{2}{c}{BERT+NMT}\\\hline
         IWSLT.FT.$tbase.bpe10k$ & 27.4 & 17.6  \\
         IWSLT.FT.$tbase.dec3.bpe10k$ & 27.2  & \textbf{18.1}  \\\hline
         Wiki.FT.$tbig.bpe32k$ & 26.9 & 17.6 \\
         Wiki.FT.$tbig.dec3.bpe32k$ & \textbf{27.7} & 17.8  \\\hline
         News.FT.$tbig.bpe32k$ & 27.1 & 17.9 \\
         News.FT.$tbig.dec3.bpe32k$ & 27.6 & 17.9 \\
         
    \end{tabular}
    \caption{IWSLT dataset results. IWSLT.FT: encoder is initialised with BERT model trained on IWSLT data; $tbase$/$tbig$: transformer base/big architecture for NMT model; $dec3$: decoder layers reduced for 6 to 3; $bpe10k$/$bpe32k$ : amount of BPE merge operations used for source language, learnt on the same dataset as BERT model (IWSLT or Wiki+News). }
    \label{tab:res_iwslt}
\end{table}
Table \ref{tab:res_iwslt} reports the results of different sets of the experiments on IWSLT data. First, we observe that BERT pretrained model improves over the baseline, in any settings (BPE vocabulary, model architecture, dataset used for pretraining). In particular, it is interesting to mention  that without pretraining, both $tbig.bpe32k$ and $tbig.bpe10k$  models diverge when trained on IWSLT. However, BERT pretraining gives a better initialization point, and allows to achieve very good performance both for en-de and en-ru. Thus, such pretraining can be an interesting technique in low-resource scenarios.    

We do not observe big difference between IWSLT pretrained model and News/Wiki pretrained model. We therefore may assume that News/Wiki BERT model can be considered as "general" English pretrained encoder, and be used as a good starting point in any new model translating from English (no matter target language or domain).

\section{Discussion}\label{sec:discussion}

BERT pretraining has been very successful in NLP.
With respect to MT, it was shown to provide better performance in \newcite{XLM2019,EdunovEtAl2019} and allows to integrate large source monolingual data in NMT model as opposed to target monolingual data usually used for backtranslation.

In this experimental study, we have shown that:
\begin{itemize}
    \item The next sentence prediction task in BERT is not necessary to improve performance - a \textit{masked LM task} already is beneficial.
    \item It is beneficial to train BERT on the \textbf{source corpora}, therefore supporting the claim that pretraining the encoder provide a better intialization for NMT encoders.
    \item Similar to \newcite{EdunovEtAl2019}, we observe that the impact of BERT pretraining is more important as the size of the training data decreases (WMT vs IWSLT).
    
    \item Information encoded by BERT is not sufficient to perform the translation: NMT encoder encodes both information specific to source sentence, and to the target sentence as well (cf the low performance of BERT frozen encoder). 
    \item Pretraining the encoder enables us to train bigger models. In IWSLT, the transformer big models were diverging, but when the encoder is initialized with pretrained BERT the training became possible. For WMT14, training a 12 layer encoder from scratch was problematic, but News.Emb model (which contains 12 encoder layers) was trained and gave one of the best performances on WMT14.
    \item Finetuning BERT pretrained encoder is more convenient : it leads to similar performance compared to reusing BERT as embedding layers, with faster decoding speed.
    
    \item BERT pretrained models seem to be generally better on different noise and domain test sets. However, we didn't manage to obtain clear evidence that these models are more robust. 
  
\end{itemize}

This experimental study was limited to a particular dataset, language pair and model architecture. However, many other combinations are possible. First, similar type of study needs to be performed with BERT pretrained model for NMT decoder. Also, the model can be extended to other scenarios with BERT models such as \newcite{cloze2019fair}. In addition, the comparison with ELMO embeddings is also interesting as in \newcite{EdunovEtAl2019}. Using embedding mostly influenced by neighboring words seems to echo the recent results of convolutional self attention network \cite{csan}. Using  convolutional self attention network in BERT could bring additional benefit for the pretrained representations. Another direction could look at the impact of the number of layers in BERT for NMT.

Besides, one key question in this study was about the role of encoder in NMT as the roles of encoders and decoders are not clearly understood in current neural architectures. In the transformer architecture, the encoder probably computes some interlingual representations. In fact, nothing constraints the model in reconstructing or predicting anything about the source sentences.
If that is the case, why would a monolingual encoder help for the NMT task?

One hypothesis is that encoders have a role of self encoding the sentences but also a translation effect by producing interlingual representations. In this case, a monolingual encoder could be a better starting point and could be seen as a regularizer of the whole encoders.
Another hypothesis is that the regularization of transformers models is not really effective and simply using BERT models achieve this effect.

\section{Conclusion}
In this paper, we have compared different ways to use BERT language models for machine translation. In particular, we have argued that the benefit of using pretrained representations should not only be assessed in terms of BLEU score for the in-domain data but also in terms of generalization to new domains and in terms of robustness.

Our experiments show that fine-tuning the encoder  leads to comparable results as reusing the encoder as an additional embedding layers. However, the former has an advantage of keeping the same model size as in standard NMT settings, while the latter adds additional  parameters to the NMT model which increases significantly the model size and might be critical in certain scenarios. 

For MT practioners, using BERT has also several practical advantages beyond  BLEU score. BERT can be trained for one source language and further reused for several translation pairs, thus providing a better initialization point for the models and allowing for better performance. 

With respect to robustness tests, the conclusion are less clear. Even if pretrained BERT models obtained better performance on noisy test sets, it seems that they are not more robust than expected and that the potential of masked LM tasks is not fully exploited for machine translation. An interesting future work will be to assess the robustness of models from \newcite{song2019mass}.


\bibliographystyle{acl_natbib}
\bibliography{references}

\end{document}